\documentclass[graybox]{svmult/styles/svmult}

\usepackage{mathptmx}       
\usepackage{helvet}         
\usepackage{courier}        
\usepackage{type1cm}        
\usepackage{makeidx}         
\usepackage{graphicx}        

\usepackage{multicol}        
\usepackage[bottom]{footmisc}
\usepackage{courier}
\usepackage{url}
\usepackage{amsmath}
\usepackage{amsfonts}
\usepackage{hyperref}

\usepackage[textfont=footnotesize]{subfig}

\usepackage{color}
\usepackage{soul}
\definecolor{orange}{rgb}{0.93, 0.53, 0.18}

\makeindex             

\begin{document}

\title*{iCub}
\author{Lorenzo Natale, Chiara Bartolozzi, Francesco Nori, Giulio Sandini, Giorgio Metta}
\institute{Lorenzo Natale, Chiara Bartolozzi, Francesco Nori, Giulio Sandini, Giorgio Metta \at Istituto Italiano di Tecnologia, via Morego 30, 16163, Genova, Italy, \email{name.surname@iit.it}}
%
%
\maketitle

This is a post-peer-review, pre-copyedit version of an article published in Humanoid Robotics: A Reference, Springer. The final authenticated version is available online at: \href{https://doi.org/10.1007/978-94-007-6046-2\_21}{https://doi.org/10.1007/978-94-007-6046-2\_21}
\\
\\
Cite this Chapter as:\\
Natale L., Bartolozzi C., Nori F., Sandini G., Metta G. (2017) iCub. In: Goswami A., Vadakkepat P. (eds) Humanoid Robotics: A Reference. Springer, Dordrecht. https://doi.org/10.1007/978-94-007-6046-2\_21
\\
\\

\abstract*{Each chapter should be preceded by an abstract (10--15 lines long) that summarizes the content. The abstract will appear \textit{online} at \url{www.SpringerLink.com} and be available with unrestricted access. This allows unregistered users to read the abstract as a teaser for the complete chapter. As a general rule the abstracts will not appear in the printed version of your book unless it is the style of your particular book or that of the series to which your book belongs.
Please use the 'starred' version of the new Springer \texttt{abstract} command for typesetting the text of the online abstracts (cf. source file of this chapter template \texttt{abstract}) and include them with the source files of your manuscript. Use the plain \texttt{abstract} command if the abstract is also to appear in the printed version of the book.}

\abstract{In this chapter we describe the history and evolution of the iCub humanoid platform. We start by describing the first version as it was designed during the RobotCub EU project and illustrate how it evolved to become the platform that is adopted by more than 30 laboratories world--wide. We complete the chapter by illustrating some of the research activities that are currently carried out on the iCub robot, i.e. visual perception, event-driven sensing and dynamic control. We conclude the Chapter with a discussion of the lessons we learned and a preview of the upcoming next release of the robot, iCub 3.0.}

\newpage

\section{Introduction}
\label{sec:1}

Robotics has been growing at constant pace, with the expectation that robots will find application outside research laboratories. Within this field humanoid robots are being studied because they offer great advantages in terms of flexibility and versatility. In addition, the anthropomorphic embodiment is required for natural human-robot interaction and to model human behavior. Many humanoid platforms have been built so far: their application domain ranges from service robots~\cite{armar,asimo,romeo}, aerospace~\cite{justin, robonaut, valkyrie}, entertainment~\cite{nao, hrp4} and the civil domain~\cite{valkyrie,hrp3,hubo,negrello16}. Other examples are humanoid robots that have been developed primarily as research platforms to study cognition~\cite{Metta20101125}, locomotion~\cite{hrp3,hubo,tsagarakis2013compliant} or human robot interaction~\cite{nao,simon}. This robot `zoo' demonstrates great variability in the choice of the kinematics, type of actuation, materials and sensors. None of these platforms, unfortunately, has reached wide adoption (the sole exception is the Nao robot~\cite{nao}, which, however, has been adopted mainly by the Human Robot Interaction community). This is certainly due to the high cost of the platforms, especially those that are more sophisticated. Another reason is that none of the available platforms has enough functionalities or has reached enough maturity to become a standard platform. The recent progress in AI -- mainly machine vision and learning -- raises further the demand for a robotic platform that can be used off-the-shelf either for research or for developing applications.

iCub is a humanoid platform that was developed for research~\cite{Metta20101125}, within the RobotCub project, a 5 year effort financed by the European Union 6\textsuperscript{th} Framework Program to study cognition in humans and artificial systems. It was designed to be used as a testbed for algorithms and theories modeling aspects of human cognition, including learning, perception and motor control. The RobotCub project also aimed at fostering the creation of a community that could experiment on the same platform, thus speeding up progress in two ways. Firstly, by providing researchers with a ready-to-use, complete, humanoid platform, with sophisticated kinematics, a human-like sensory system, and -- importantly -- a mature documentation and software API. Secondly, by facilitating code exchange among researchers working in different laboratories, thus promoting code re-use and benchmarking. In the past years many research groups have joined the iCub community: as of today 35 robots have been built and they are hosted by approximately the same number of laboratories located in Europe and world-wide.

The iCub platform has grown at a steady pace both in terms of mechatronics and software capabilities. In this Chapter we describe the history and evolution of the platform during the past 12 years of research. We start from the first version developed during the RobotCub EU project (i.e. iCub 1.0) and we continue by describing how it evolved in subsequent revisions (which is also called iCub 2.5). We complement the Chapter with an overview of the research activities that are being carried out at the iCub Facility at the Istituto Italiano di Tecnologia, with the aim of advancing the capabilities and autonomy of the robot. We conclude the Chapter by discussing the lessons learned and by providing a preview of the design choices of the upcoming iCub 3.0.

\begin{figure}[h]
\centering
\includegraphics[width=6cm]{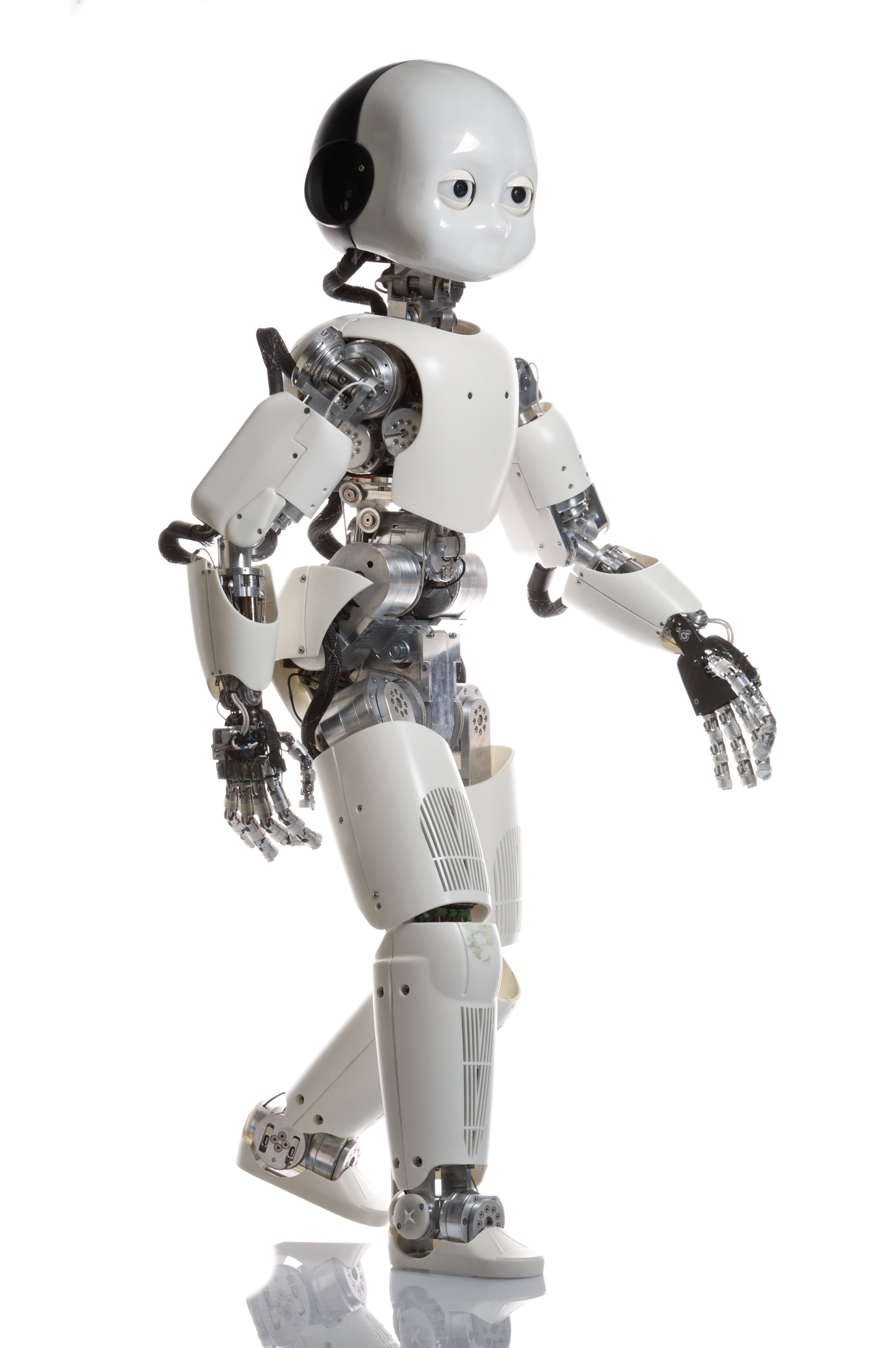}
 \caption{A picture of the iCub.}
    \label{fig:icub-kinematics}
\end{figure}

\section{iCub 1.0}
\label{sec:icub10}
The design of the iCub replicates the proportions of a three-year-old child in a fully-fledged humanoid robot which includes legs, torso, arms and head.  The design focused on giving the robot the ability to manipulate objects with locomotion and a certain dexterity. To simlify the control of the robot, it was decided to focus on crawling rather that walking: for this reason the legs and arms were designed to support crawling on all fours.

\subsection{Mechanical design}
To obtain correct proportions, the size of the robot closely follows data from anthropomorphic tables~\cite{tilley2002}, in particular considering the proportions of the limbs of a three-year-old child. Body simulations allowed to determine the kinematic features of the human body that needed to be replicated to perform a set of desired tasks and motions~\cite{tsagarakis2007}. The resulting kinematics is reported in Figure~\ref{fig:icub-kinematics}. The arms have 7 Degrees of Freedom (DoF) each to increase the robot dexterity and the reachable workspace. The latter is further extended thanks to the mobility of the waist, which is actuated by three motors (namely pitch, roll and yaw). The hands of the robot~\cite{parmiggiani12,schmitz10} were designed to closely resemble human hands and for this reason they have been equipped with 5 fingers. Four joints are in the thumb, while the index, middle, ring and little finger have three joints each. Joints at the base of the index, ring and little fingers provide an abduction-adduction mechanism. In total the hand has 19 joints\footnote{We do not count the three DoF of the wrist.}, that, for practical reasons, are (under) actuated by 9 motors. Four motors directly actuate the proximal and middle phalanges of the index and middle index, while three motors control the rotation, proximal and distal phalanges of the thumb. One motor controls the coupled motion of the phalanges of the ring and little fingers, and another motor controls the abduction/adduction of the fingers. The proximal and middle phalanges of the index and middle fingers are coupled with tendons.  Encoders provide feedback on the position of the motors and all the 19 joints, so to measure the configuration of the hand. The hand is extremely compact, it measures approximatly 150~mm in length, 60~mm in width and 25~mm in thickness.

\begin{figure}[b]
\centering
\includegraphics[width=6cm]{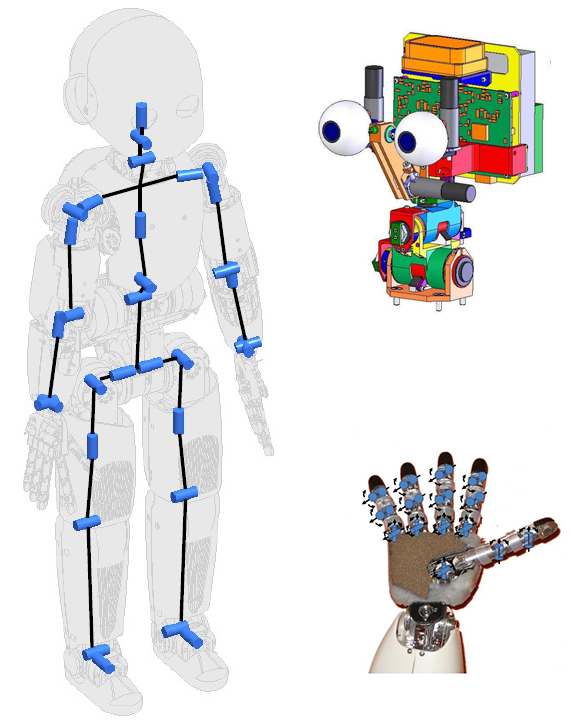}
 \caption{A picture representing the kinematics of the iCub 1.0 (left). The degrees of freedom of the eyes and hands are reported to the right. Notice that the abduction is represented separately for each finger but it corresponds, in fact, to a single DoF.}
    \label{fig:icub-kinematics}
\end{figure}

In the iCub all joints are actuated with rotary electrical motors, because the maturity of this technology allows achieving higher robustness and reliability. To increase the level of integration and remove the extra space required by the housing, frameless motors are used to actuate the joints that required higher torques, like the waist, legs and shoulder~\footnote{Kollmorgen-DanaherMotion RBE motors with CSD Harmonic Drive}. Brushless motors were preferred over DC motors because they offer higher torque, whereas Harmonic Driver reducers were chosen because they have almost no backlash and high reduction ratio. The remaining joints (namely the joints of the head, forearm and hands) use DC motors, instead.

\begin{figure}[h]
\centering
\includegraphics[width=6cm]{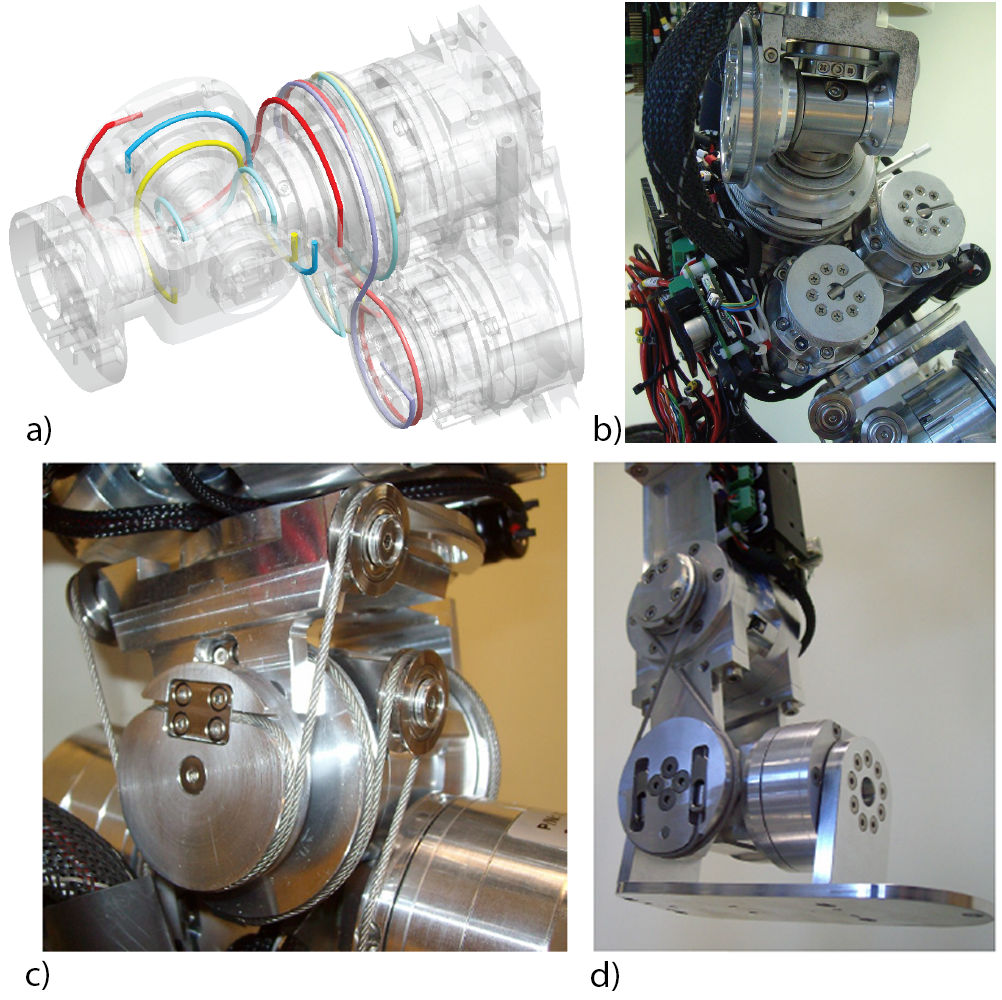}
 \caption{Details of the actuation of shoulder (a) and (b), waist (c) and ankle (d) joints.}
    \label{fig:icub-actuation}
\end{figure}

The iCub 1.0 widely adopts cable drive transmission using steel tendons. This allows transmitting motion with high efficiency, routing the power between bodies rotating along different axis using various types of pulleys. In turn this allows driving distal joints using motors located close to the body, and consequently reducing the inertias of the links and the torque required to drive them. A notable example is the actuation of the three DoF of the shoulder -- which are actuated with motors mounted inside the torso -- and the hand joints -- for which 7 of the 9 motors are mounted on the forearm.
%
Other joints actuated with cable driven transmission are: the elbow, the torso, the hips and the ankles. Figure~\ref{fig:icub-actuation} shows how cables are used in the shoulder, waist and ankle joints. To maintain appropriate tension of the tendons and reduce backlash, tensioners are mounted either at the extremity (single screw) or along the cables (double screws). 

One of the design specification of the iCub was its weight, which was decided to be approximately 20-25~Kg. To maintain the total weight of the robot within this range it was important to reduce the mass of the structural elements. The majority of the parts are fabricated with aluminium alloys (i.e. Al6082 and Al7075, the latter with higher fatigue strength was adopted for joints that required mode demanding mechanical properties). Joint shafts -- which undergo high mechanical stress -- are built using high resistance stainless steel (i.e. 39NiCrMo3). More details about the kinematics and the actuation system of the iCub can be found in~\cite{parmiggiani12}.

The head of the iCub 1.0 is depicted in Figure~\ref{fig:icub-kinematics}. It has three DoF neck providing pitch, roll and yaw. The eyes are two cameras which are actuated by three motors that approximate a human-like oculomotor system with common tilt, version and vergence for tracking targets in space. The atuation system of the head is made with DC motors. The latter are directly connected to the neck joints, while an actuation system with toothed belts is used to transmit motion to the eyes.

\subsection{Sensors and electronics}

The iCub 1.0 is equipped with a set of embedded microcontrollers for motor control, analog to digital conversion and a PC104 card hosting an Intel Core 2Duo 2.16~MHz Pentium processor with 1~GB of RAM. These boards are interconnected using several CAN bus lines each with the maximum bandwidth of 1~Mbps. To distribute the communication load different CAN bus lines are routed from the PC104 to the arms, the torso and the legs. In total ten CAN bus lines are used to support communication with the control boards and the sensors distributed on the robot. Besides the CAN bus interface the PC104 is equipped with Gigabit ethernet and firewire bus. The firewire is used to read images from the cameras. The Gigabit ethernet interfaces the robot with a distributed cluster of computers performing demanding computation (like visual processing, machine learning and whole-body control). The PC104 works mostly as a two-directional hub providing communication between the software on the cluster and the embedded system. In addition, it runs time critical control loops like joint torque computation and reaching in the Cartesian space~\cite{pattacini10}. The PC104 also mounts an inertial unit from XSense measuring three axis linear acceleration and angular velocities and microphones for auditory input.

Two sets of boards were designed to control Brushless and DC motors. These boards can perform position, velocity and torque control by modulating a PWM signal at the maximum power of 20~A at 48~V (for brushless motors) and 1~A at 12~V for DC motors. The electronic boards were custom-made for better integration in the available space close to the motors. Position feedback is available using incremental and absolute encoders at the motors.

The iCub 1.0 mounts six-axis force-torque sensors~\cite{tsagarakis2007} in the arms and legs. Each sensor integrates a custom electronic board for data acquisition and signal conditioning. The board is able to perform analog to digital conversion, digital filtering and the linear transformation required to convert the signals from the strain gauges into 3 axis torques and forces. The board has a CAN interface over which it can broadcast the acquired values at the maximum rate of 1~KHz.

Finally, a particular effort was devoted to the design and implementation of a distributed tactile system, illustrated in Section~\ref{sec:tactile-system}.

\section{iCub 2.5}
\label{sec:icub2}

The iCub community started growing during the RobotCub project, when we built several copies of iCub 1.0\footnote{Within the RobotCub projet we built 10 copies of the robot, 3 of these were for partners of the project whereas 7 were distributed to winners of an open call. In addition, 7 robots were built for the FP7 EU projects ITALK and IM-CLeVeR.}. The use of the robot in different contexts within the community allowed for a fast testing phase that led to the design of various improvements which are today integrated in the iCub 2.5. The main improvements addressed the head, with new actuation, enhanced facial expressions and audio system, the legs supporting balancing and locomotion and new electronics.

\begin{figure}[b]
\centering
\includegraphics[width=8cm]{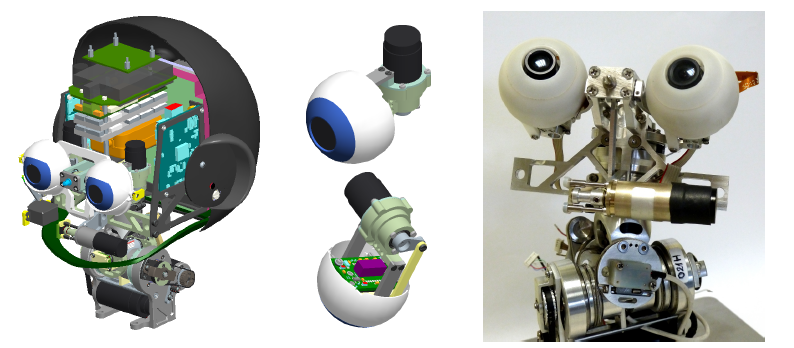}
 \caption{CAD drawings of the iCub 2.5 head (left) with details of the eyes (center). A picture of the head without electronics (right).}
    \label{fig:icub-head2}
\end{figure}

Some of the problems affecting iCub 1.0 are in the head. The motor of the neck actuating the pitch joint are underengineered  and, at times, overheat or do not provide enough torque. Another problem is a significant backlash in the eye movement which introduces problems for stereo vision and large errors when triangulating the gaze position. To address these problems a new revision of the head was designed with modified neck and eyes systems. The new head assembly employs an epicyclic transmission~\cite{brooks99,aryananda04}, it reduces the weight by approximately 1~Kg and it has 8~Nm peak torque on the pitch and roll axes with a three-folds increase of the output torque delivered to the joints~\cite{parmiggiani12}. Backlash in the actuation of the eyes is removed using zero-backlash Harmonic Drive speed reducers (a similar solution was proposed by~\cite{asfour08}). Figure~\ref{fig:icub-head2} shows the CAD drawings and a picture of the new head.

Facial expressions in the first version of the iCub head are too simple and badly suited for human-robot interaction. The eyelids are realized using two parts which covered the eyes from above and below meeting half-way at the level of the camera. This produced unnatural expressions, because in humans eyelids achieve complete closure with the upper eyelid only. In addition, facial expressions are rendered using a matrix of LEDs mounted behind the semi-transparent face cover. This makes the original iCub head insufficient to display complex features of human speech movement. Finally, the iCub 1.0 does not have a speaker and speech synthesis is commonly done using a speaker connected to an external computer. A more sophisticated system was designed in collaboration with the GISPA Lab, CNRS Grenoble~\cite{parmiggiani15}. This new system can be optionally mounted on the iCub head. It integrates a moving jaw, mechanically actuated lips, a loudspeaker close to the mouth, and eyelids modified to achieve a more human-like movement. Finally, the performance of the audio system was improved by using high-quality microphones and reducing the noise generated by the cooling fans of the CPU\footnote{This feature is actually optional, because it uses an Intel\textregistered Atom\textsuperscript{TM} D525 which delivers less performance than the standard Intel\textregistered Core 2 Duo 2,16 Ghz.}. Overall these enhancements require 6 additional motors to actuate the DoF of the jaw (one motor), lips (4 motors) and eyelids (one motor).

\begin{figure}[t]
\centering
\includegraphics[width=10cm]{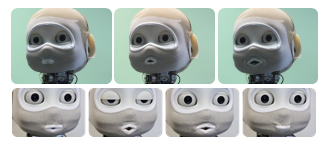}
 \caption{Bottom: Facial expressions, lips and eyelids motion. Top: some facial expressions achieved by reproducing data measured from a human speaker. Images adapted from~\cite{parmiggiani15}.}
    \label{fig:icub-talking}
\end{figure}

To cover the actuators of the jaw and lips we use a ``skin'' made of stretchable fabric (i.e. Lycra). The cover has openings for the eyes and the mouth, and, to allow ventilation, on the back.  Lips are made with hemlines obtained by sewing extra layers of Lycra. The corners of the lips are attached to the actuation system of the lips. Figure~\ref{fig:icub-talking} shows some of the facial expressions that can be achieved with the head.

\begin{figure}[b]
\centering
\includegraphics[width=8cm]{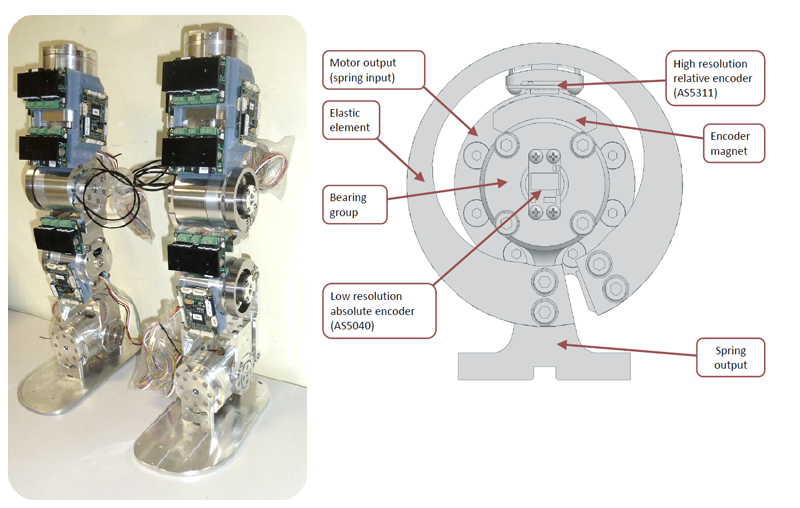}
 \caption{Left: a picture of the legs. Right: front view of the elastic module and its main components. Images adapted from~\cite{parmiggiani12b}.}
    \label{fig:icub-legs2}
\end{figure}

As stated in Section~\ref{sec:icub10} the legs of the iCub were originally developed to support crawling on all four. Although the motors are strong enough to support the weight of the robot, balancing and locomotion are made difficult by the small size of the feet, the relatively range of motion of the ankle and the absence of accurate torque feedback. These issues have been addressed in a new revision of the legs. The new legs adopt some of the modifications already introduced in the COMAN robot~\cite{tsagarakis11}. The major novelty was the design of a series elastic actuator (SEA~\cite{pratt95}) for the knee and ankle joints.
Since their invention~\cite{pratt95} SEAs have been widely adopted in humanoid robots~\cite{brooks99,edsinger04,pratt08}. Rotational SEAs are integrated in the COMAN robot to achieve compliance and torque feedback. One of the problem we had to solve was how to obtain springs whose stiffness was large enough to support the weight of the robot and avoiding maximum compression. To this aim we introduced a ``curved beam'' between the motor shaft and the link that deflects when torque that is applied at the extremities. This beam acts as a ``C-shaped'' spring, whose deflection is measured with a high resolution magnetic encoder. The SEA element and its main components are shown in Figure~\ref{fig:icub-legs2} (right). The new legs also include a 6 axis Force/Torque sensor placed between the foot and the ankle and a set of tactile sensors on the sole-foot. The transmission mechanism of the ankle was also modified to remove the cable drive which introduces, in the original design, unwanted elasticity. The new transmission adopts a four bar linkage already tested on the COMAN robot. Figure~\ref{fig:icub-legs2} (left) shows a picture of the legs before they are mounted on the robot.

\subsection{Sensors and electronics}

\begin{figure}[h]
\centering
\includegraphics[width=12cm]{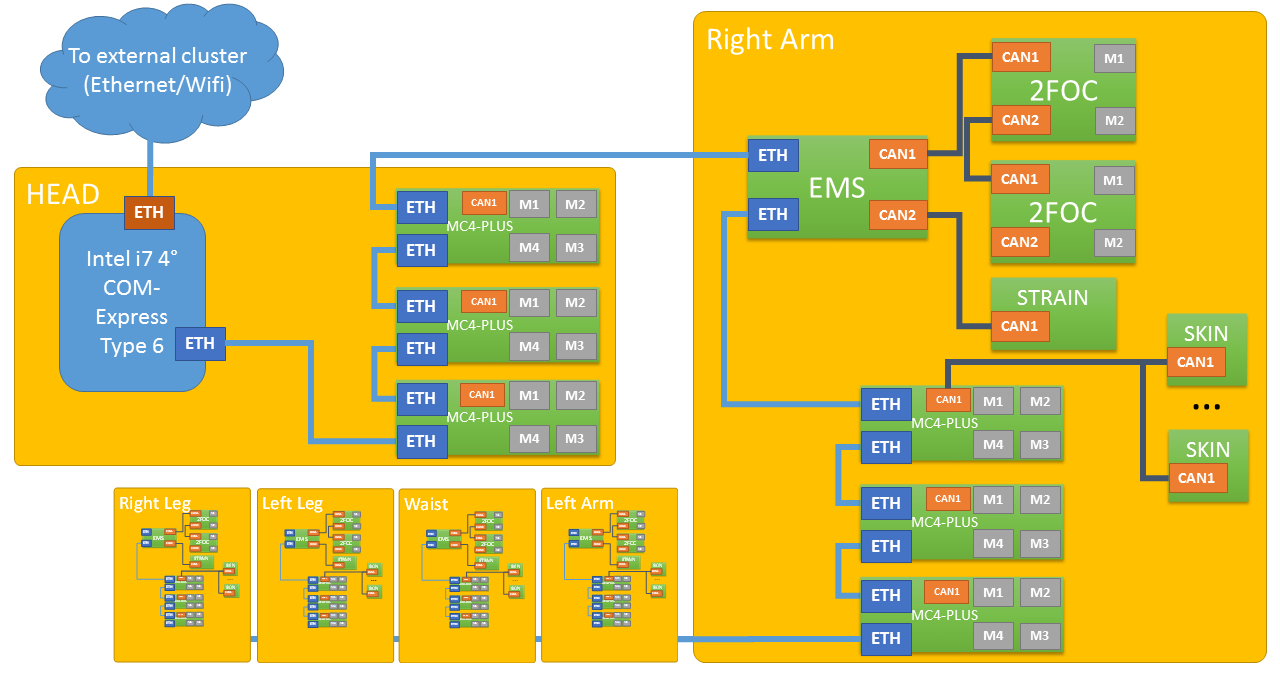}
 \caption{A schematic representation of the network of embedded boards on the iCub 2.5. The picture shows the arrangement of the boards that control the motors on the head (up to 12 motors) and the right arm (16 motors). The right arm also hosts the boards that read the F/T sensor (STRAIN) and the tactile system (SKIN). See text for details.}
    \label{fig:electronics}
\end{figure}

The control boards in the iCub 2.5 have been redesigned completely. This redesign was primarily aimed at supporting current control in the brushless motors and increasing the available bandwidth on the embedded network. The architecture is illustrated in Figure~\ref{fig:electronics}. We can identify two subnetworks: ethernet and CAN bus. The ethernet network is the main backbone, while the CAN bus provides local board-to-board connectivity. The ethernet network is made by boards that mount 32~bits ARM Cortex-M4 processors, and are equipped with CAN bus ports, analog to digital converters and an inertial unit. An ethernet switch with two ports allows connecting the boards in daisy-chain and -- using a custom protocol -- exchange packets with the PC on the head of the robot at the frequency of 1~Khz. These boards control up to 4 DC motors directly (MC4-PLUS) or work as a bridge to the low-level layer using CAN bus (EMS). The low-level boards include boards for controlling up to 2 Brushless motors using Field Oriented Control (2FOC), reading F/T sensors (STRAIN), or the tactile sensors (SKIN). The PC on the head is an Intel i7 COM-Express that works as a hub interconnecting (wired or wireless) the embedded system on the robot with the external cluster of computers.

\section{The Software Infrastructure}
\label{sec:software}

The evolution of the software infrastructure has been described in details in~\cite{natale16}, therefore only a brief description is provided here. The software on the iCub is based on the YARP middleware which provides a variant of the publish-subscribe framework inspired by the observer design pattern~\cite{themanyfacesEugster03}. The YARP middleware is similar to ROS~\cite{quigley09} and it is fair to say that the two projects share the same philosophy. With respect to other middleware YARP provides a plug-in system for device drivers and protocols. YARP is therefore a flexible middleware that can easily interoperate with other frameworks. One of the recent efforts has been the development of protocols providing compatibility with ROS, giving the iCub users access to the large amount of software developed by the ROS community. In recent work we have also extended the communication layer to give users the possibility to assign different priorities to the communication channels in the network~\cite{paikan15besteffort}. We are currently working to implement tools for controlling the priorities depending on the task at hand, so that messages from non-critical components (like video streams for debugging and visualization) do not interfere with messages exchanged within control loops.

An important feature in YARP is the robot abstraction layer that separates user code from robot specific code. This abstraction layer allowed to gradually expose new functionalities as they became available in the hardware platform, with minimal impact to the user code. Thanks to this abstraction layer various iCub simulators have been implemented using different engines~\cite{tikhanoff08,mingo14,habra15}. All these simulators share the same robot interface and can be used interchangeably without modification to the code.

The iCub software repositories are maintained in the ``robotology'' organization in github\footnote{https://github.com/robotology}. The organization received contributions from 243 developers, of which 160 have contributed in the last 12 months\footnote{source: www.openhub.net}. To improve the quality of the core software of the robot (namely the firmware, the drivers and the YARP libraries) we have recently adopted code reviews using the functionalities provided by github. We have also adopted a framework for test driven development of robot software~\cite{paikan15rtf}. The suite of tests for the iCub\footnote{https://github.com/robotology/icub-tests} is currently used to test various aspects of the software, such as strict compliance to the specifications, correct configuration and specific bugs (regression tests).

\section{Tactile system}
\label{sec:tactile-system}

\begin{figure}[b]
\centering
\includegraphics[width=9cm]{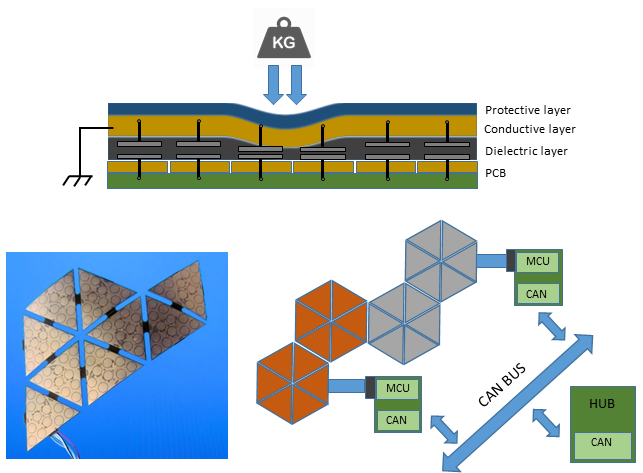}
 \caption{The iCub skin. Top: a schematic representation of the three layers that form the set of capacitors each providing pressure information. Bottom: details of the triangular elements and how they are interconnected to form a mesh of sensors that can be read using CAN bus interface.}.
    \label{fig:skin-triangles}
\end{figure}

\begin{figure}[b]
\centering
\includegraphics[width=9cm]{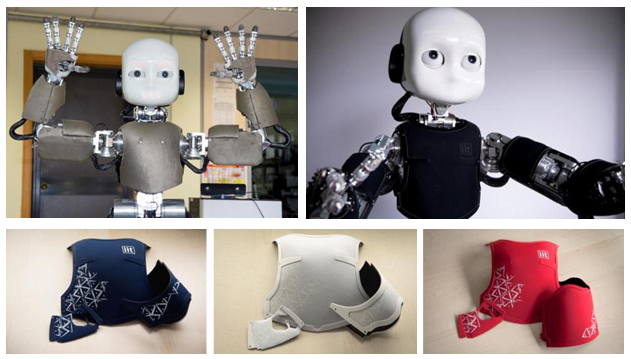}
 \caption{The first version of the skin used a silicone layerd as dielectric, covered with conductive lycra (top-left). In the second version of the skin these layers were replaced by a sandwitch of three layers made of fabrics glued with industrial techniques. Among the advantages of this solution is the fact that the production of the fabric is automated and more reliable. The picture on the top-right shows the the latest version of the skin. The figures on the bottom show possible customizations in different colors.}.
    \label{fig:skin-evolution}
\end{figure}

\begin{figure}[b]
\centering
\includegraphics[width=8cm]{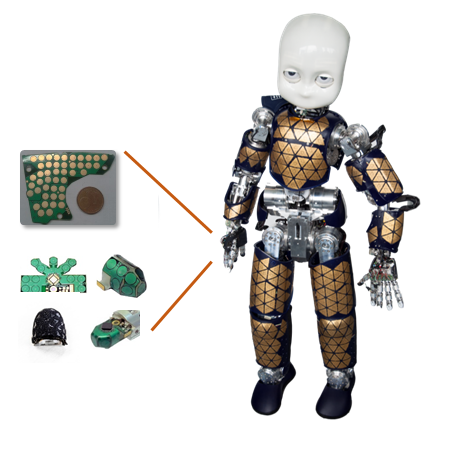}
 \caption{The picture illustrates the placement of the tactile sensors on the body of the robot (right), including the palms and fingertips (left). The iCub is equipped with 4488 sensors: 104 in each hand (12 on each fingertips and 44 in the palms), 230 and 380 respectively in each forearm and arm, 440 in the torso and 1310 in each leg (including the feet -- not shown in this picture). }.
    \label{fig:skin-placement}
\end{figure}

\begin{figure}[b]
\centering
\includegraphics[width=12cm]{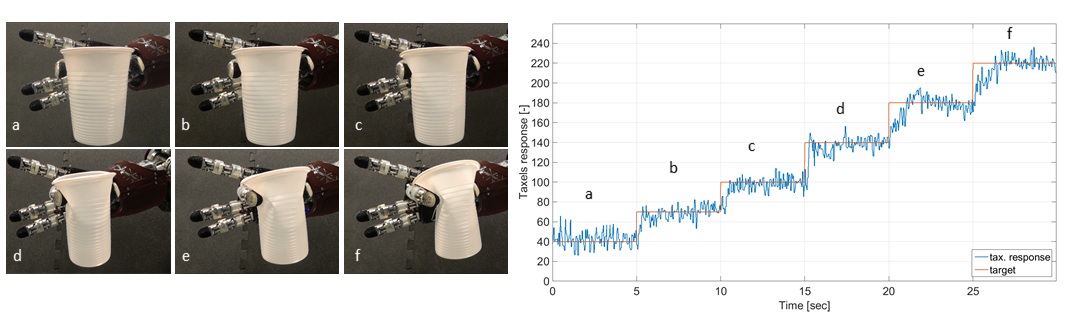}
 \caption{This experiment shows control of grip of a fragile object (in this case a plastic cup). The controllers regulate the motor PWM to keep a certain values of pressure (in this case measured as the average of all 12 sensors on the index fingertip). The desired value of pressure is increased in steps. The plot on the right shows the value of pressure measured by the fingertip, while the frames on the left shows how the object deform as the grip strength increases (details are in~\cite{regoli2016}).}.
    \label{fig:plastic-cup}
\end{figure}

Most of the iCub body is covered with a system of capacitive tactile elements. This ``skin'' is obtained by gluing various layers which form a set of capacitors each providing pressure information (Figure~\ref{fig:skin-triangles}, top). The bottom layer consists in a printed circuit board (PCB) which provides the bottom plates of the capacitors. The second layer is a dielectric material which is soft and deforms when pressure is applied on the sensor. The remaining part of the system is made by a conductive cover which forms the top plates of the capacitors and provides a protective coat. The basic unit of the system is a triangle which contains 10 sensors/capacitors (Figure~\ref{fig:skin-triangles}, bottom). Triangles are interconnected in a deformable mesh which conforms to the surface to be covered.  To reduce the number of wires, the triangles are interconnected through a serial digital bus: capacitance values are read by a microcontroller units (MCU) and travel across triangles to a hub micro-controller. The hub is equipped with a CAN bus interface which broadcast all values to a single interface board. Figure~\ref{fig:skin-triangles} illustrates how triangles are interconnected to form a ``skin patch'' while Figure~\ref{fig:electronics} describes how each ``patch'' is integrated in the embedded system on board of the robot.

Figure~\ref{fig:skin-evolution} shows the evolution of the tactile system: from a prototype made by layering silicone foam and conductive lycra~\cite{schmitz2011} to an industry-grade implementation that uses customizable fabric~\cite{maiolino2013flexible}. This technology has been adopted for the triangular elements that cover the larger parts of the body (arms, torso and legs). It has also been customized to cover the palm of the robot (for which a custom PCB has been designed) and the fingertips. In the latter case, the problem was more difficult due to the minimal space available and large curvature of the surface of the fingertip. The solution we adopted was to design a custom PCB with 12 sensors, which is wrapped around a plastic support. The fingertip is covered with the three layers of fabric that complete the sensor. The first version of the fingertip used silicone foam as a dielectric material covered with carbon black~\cite{schmitz10RoMan,schmitz2011}. The silicone foam has been recently substituted with fabric, leading to larger sensitivity, lower hysteresis and mechanical robustness~\cite{jamaliANewDesign}.

In its latest versions the iCub has been covered with a toal of 4488 sensors. Figure~\ref{fig:skin-placement} shows how the sensors are distributed on the body of the robot (notice that the picture does not show the sensors on the feet). The response of the tactile system has been extensively characterized by measuring, sensitivity, repeatability, spatial sensitivity and hysteresis. Details of the experimental validation of the tactile system are in~\cite{maiolino2013flexible} for the triangles and in~\cite{jamaliANewDesign} for the fingertips. The tactile sensors on the fingertips have been used for grip control and object re-grasping~\cite{regoli2016}. Figure~\ref{fig:plastic-cup} shows an experiment in which the robot controls the pressure exerted by the fingers on a fragile object.

%
%


\begin{figure}
  \centering
    \includegraphics[width=.9\textwidth]{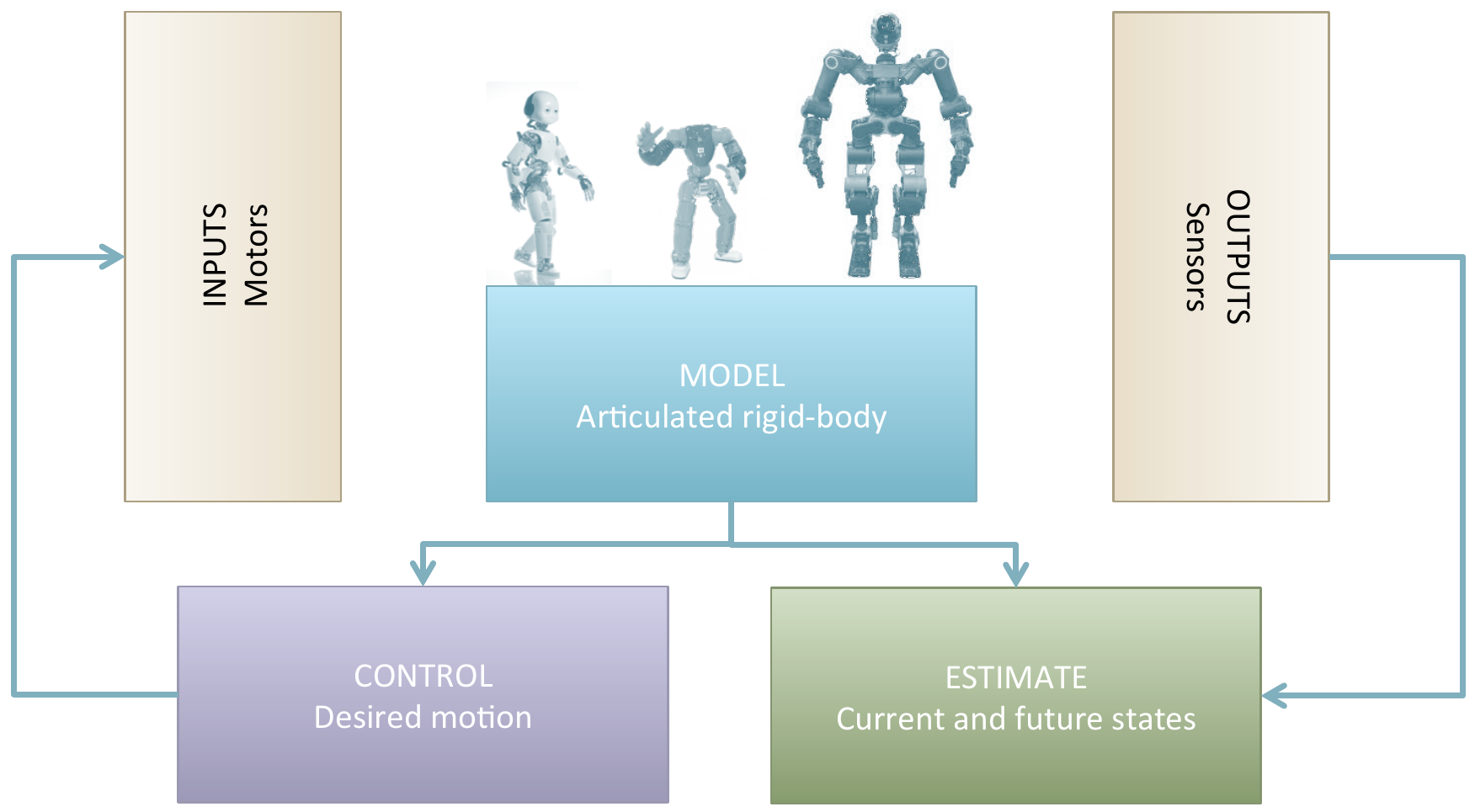}
  \caption{Block diagram of the whole-body control framework. The entire
  framework is structured around the \texttt{wholeBodyInterface}\protect\footnotemark, a software
  abstraction layer for simplifying code use across different robots
  (in the image the robots iCub, COMAN and WALKMAN for which the \texttt{wholeBodyInterface}
  is already available).}
  \label{fig:figures_wbd_wbd_structure}
\end{figure}
\footnotetext{\url{https://github.com/robotology/wholebodyinterface}.}

\section{Whole-body dynamics modelling, estimation and control}
\label{sec:wbd}

Activities related to whole-body interaction control can be divided
into three sub-topics: interaction modeling, estimation and control.
Modeling deals with dynamic models of interaction, with the
idea of simultaneously representing motion and forces. Estimation
deals with the problem of accurate evaluation of quantities that
cannot be directly measured with sensors (e.g. internal forces).
Control deals with the problem of controlling relevant interaction
variables (e.g. exchanged forces).

\subsection{Whole-body dynamics modelling}
\label{sec:wbd_modeling}


The iCub control framework originally adopted a fixed-based
kinematic system and progressively
evolved to use a dynamic free-floating
articulated chain. The latter approach has the following characteristics:
\begin{enumerate}

\item Under-actuation. In a free-floating robot no direct actuation
is available for the robot base position and orientation. The same formalism can
be naturally extended to include joints with series elastic actuators like the one presented in
Section \ref{sec:icub2}.

\item Global coordinates. Since the pose includes the orientation of
the base frame, we use a global parametrization and
and a non-minimal set of coordinates (e.g. quaternions).
The robot velocity vector $\nu$ includes the twist of the
base frame\footnote{\url{
http://wiki.icub.org/codyco/dox/html/dynamics_notation.html}.}
and therefore it is not the time derivative of the joints $q$ as it is often
the case for fixed-base manipulators.

\item Motions constrained by rigid contacts. Constraints acting on the system are
represented by their first oder derivative, nominally the Jacobian of the contact points.
Contacts act on the system dynamics
through the external forces $f$ which depend linearly on the
joint torques $\tau$.


\end{enumerate}

At present, the software tools available to the iCub community
include a number of sub-libraries that support manipulation of
kinematic and dynamic models. iKin is an open source library for forward-inverse kinematics
of serial-links chains with revolute joints. The iKin library contains a reliable and
flexible set of kinematic primitives fully documented. It can be used
to deal with any serial-link chains, given  the description in
standard Denavit-Hartenberg convention. To provide the iCub with
the ability to control joint torques, the iKin library
was extended with its dynamic counterpart: iDyn.
The library also provides a
computationally efficient Newton-Euler
recursion to compute forces, moments, and joint torques given joint
position, velocities, accelerations and embedded force/torque sensors
\cite{Fumagalli2012}. Finally, with the development of the new legs system
for whole-body motion and walking (as described in Section \ref{sec:icub2}), the iCub
community necessitated models of free-floating robots.
This necessity resulted in the release of yet another
open source library, named iDynTree.
%
%
The iDynTree library offers an interface which provides a
factorization of the dynamic equation suitable for identifying
the robot dynamic parameters. Thanks to this interface, a number
of experiments for identifying the robot dynamics have been
performed. Simple parametric identification techniques
\cite{Traversaro2015} have been recently extended to semi-parametric
approaches \cite{Camoriano2016} which may be adopted when
non-modeled effects must be taken into account.

\subsection{Whole-body dynamics estimation}
\label{sec:wbd_control}
%
The development of tools for the estimation of dynamic quantities in absence of
direct measure, evolved driven by the necessity to model the robot dynamics.
At first, we addressed the problem of estimating joint-level torques from embedded
force/torque sensors \cite{Fumagalli2010,Fumagalli2012}
coupled with whole-body distributed tactile sensors \cite{schmitz2011}
 to detect contact locations.
%
The proposed estimation strategy simultaneously estimates
external contact wrenches (forces and torques) and internal joint
torques.
Recent extensions of this estimation strategy have included
the possibility of fusing multiple sensor modalities~\cite{Nori2015b}
such as accelerometers, gyroscopes and force/torque sensors.

\subsection{Whole-body dynamics control}
%
%
Once external and internal forces are estimated, force control is implemented
by two control loops: an \emph{inner torque control loop} to guarantee that each motor generates a desired
torque and an \emph{outer control loop} to guarantee a desired postural compliance and to regulate the amount of force
exchanged at the contacts.
%
%
%
%

The inner torque control loop is implemented with a model-based controller
tuned for each motor. The model assumes that the $i$-th joint's torque $\tau_i$
is proportional ($k_t$) to the
voltage $V_i$
applied to the associated motor, with the additional contribution of some viscous ($k_v$) and
Coulomb ($k_c$) friction:

\begin{footnotesize}
\begin{equation} \label{eq:motorTF}
V_i  = k_t \tau_i + (k_{vp} s(\dot{\theta}_i) + k_{vn} s(-\dot{\theta}_i)) \dot{\theta}_i + (k_{cp} s(\dot{\theta}_i) + k_{cn} s(-\dot{\theta}_i)) \mbox{sign}( \dot{\theta}_i),
\end{equation}
\end{footnotesize}

\noindent
where $ \dot{\theta}_i $ is the motor velocity, $ s(x) $ is the step function (1 for $ x>0 $, 0 otherwise) and $
\mbox{sign}(x) $ is the sign function (1 for $ x>0 $, -1 for $ x<0 $, 0 for $ x=0 $). Operationally, it was
found to be important to distinguish between positive and negative rotations as represented in the model
above. We identified the coefficients $k_t, k_{vp}, k_{vn}, k_{cp}, k_{cn}$ for each joint with
an automatic procedure (as described in~\cite{nori15}, Section 3.3.2). The motor controller uses a PID controller to track the desired torque $\tau_i^d$ with a feed-forward component given by the transmission model, i.e.:

\begin{footnotesize}
\begin{equation} \label{eq:motorControl}
V_i = k_t \left( \tau^d_i - k_p \tilde{\tau_i} - k_i \int \tilde{\tau_i} \mbox{dt} \right) + [k_{vp} s(\dot{\theta}_i) +
k_{vn} s(-\dot{\theta}_i)] \dot{\theta}_i + [k_{cp} s(\dot{\theta}_i) + k_{cn} s(-\dot{\theta}_i)] \tanh(k_s \dot{\theta}_i),
\end{equation}
\end{footnotesize}

\noindent
where $\tanh()$ is used to smooth out the sign function, $k_s$ is a user-specified parameter
that regulates the smoothing action, and $\tilde{\tau_i} = \tau_i - \tau_i^d$ is
the $i$-th torque tracking error, and $k_p,k_i > 0$ are the low-level control gains. The control objective for the torque controller consists in obtaining $\tau \simeq
\tau^d$ and therefore the controller makes the assumption that the commanded value $\tau_i^d$ is
perfectly tracked by the torque controller.
%

The outer control loop is based on a two-level control strategy which guarantees a strict
hierarchical priority between the control of the robot momentum (high priority) and the
control of a postural task (lower priority). The desired robot momentum ($\dot{H}^d$)
can be guaranteed by choosing the contact forces $f$ to satisfy the following optimization
problem:

\begin{footnotesize}
\begin{equation}
\begin{aligned}
f^{*}= \arg\min_{f} & \left \| \dot{H}^{d}-\dot{H} \right \|^{2} \\
 \text{subject to } & \dot{H}=X_f+\begin{bmatrix}
mg \\
0
\end{bmatrix} \\
 & f \text{ satisfies the friction constraints} \\
 & f \text{ satisfies the positivity of the normal forces.}
\end{aligned}
\end{equation}
\end{footnotesize}

where $H \in \mathbb R^6$ is the total robot momentum\footnote{The
robot momentum is a six dimensional vector. The first three components are the robot mass
times the robot center of mass acceleration. The second three components
are the sum of the rigid bodies angular momenta, i.e. their inertia matrices multiplied
by their angular velocities.}, $H_d \in \mathbb R^6$ is its desired value, $X$ is the
matrix composed of the adjoint matrices mapping contact points into the center of mass position
\cite{nori15}, $g$ is the gravity vector, $m$ is the total robot mass and $f$ is the
vector of the contact forces.
Once the desired contact force $f^*$ are computed solving the
optimization above, the following low priority task:

\begin{footnotesize}
\begin{equation}
\begin{aligned}
 \tau^* = \arg \min_\tau & \left\| \tau - \varphi \right\|^2 \\
	        \text{subject to } &   {M}(q)\dot{{\nu}} + {C}(q, {\nu}) {\nu} + {g}(q) =  B \tau + {J}^\top f \\
	        & {J}(q) \dot{\nu} + \dot{J}(q) {\nu} = 0\\
	        & f = f^* \\
	        &  \varphi = \text{chosen to stabilize the postural task}
\end{aligned}
\end{equation}
\end{footnotesize}

takes care of selecting the joint torques $\tau$
which generate the required $f^*$, satisfy the constrained dynamics constraints and
stabilize the system around a desired posture through a suitable choice
of the variable $\varphi$ as detailed in \cite{Nava2016}.
The desired control action is finally applied to the robot by choosing motor voltages as in Equation~\eqref{eq:motorControl}
with $\tau^*_i = \tau^d_i$.

\begin{figure}[b]
\centering
\includegraphics[width=12cm]{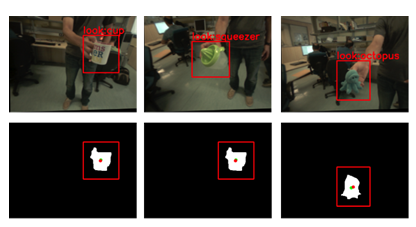}
 \caption{Object learning. An experimenter shows an object to the robot. Speech commands are used to trigger the learning session and specify the labels. Objects are segmented using disparity information and the resulting images as used as training examples.}.
    \label{fig:objects-learning}
\end{figure}

\begin{figure}[b]
\centering
\includegraphics[width=12cm]{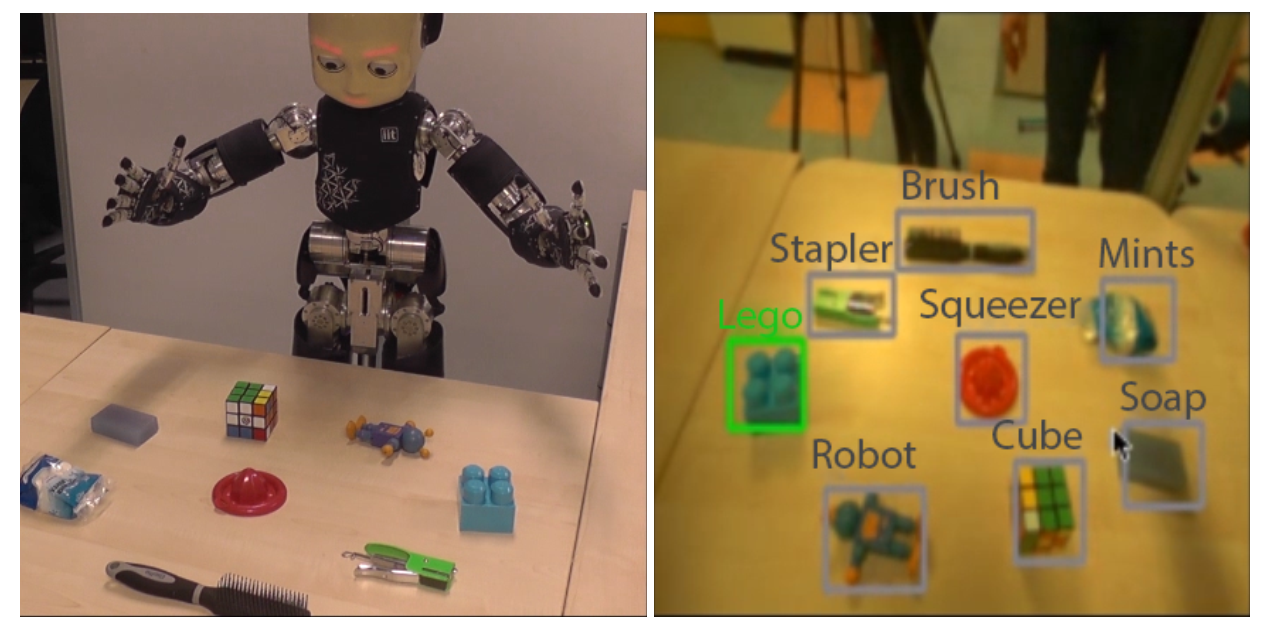}
 \caption{After learning the robot can recognize objects. In this case the system uses the features extracted from the last layer of a multi-layer convolutional neural network and a linear classifier~\cite{pasquale2016}.}.
    \label{fig:icub-objects-recognition}
\end{figure}

\section{Visual perception}
In our research we also studied topics related to low-level vision (saliency detection, motion and depth computation), object recognition and learning. One of the first problems we addressed was the implementation of a bottom-up attention system in log-polar space~\cite{rea14}. In addition, a considerable effort was devoted to the implementation of algorithms for computing visual motion and differentiate motion induced by the robot own movement (ego-motion) from the motion of the objects in the scene (independent motion detection). This is an important problem in robotics because the use of visual cues for object tracking or segmentation is made difficult by the presence of motion induced by the robot own movement. In~\cite{ciliberto11} the robot detects independent motion as those points in the image in which computation of Lucas-Kanade algorithm for optical flow fails, because of occlusions, local rotation or large speed. Under the assumption that the robot ego-motion has negligible translation component, these points belong to objects moving independently in the scene. In subsequent work the robot learns a model that allows predicting ego-motion from the speed of the encoders~\cite{ciliberto12imd} and data from the inertial unit~\cite{kumarObject2015}. This model is used to identify motion in the scene that is generated by the robot own movement and perform object segmentation~\cite{fanello2013weakly, kumarObject2015}.

For depth perception we have relied on open-source algorithms available in OpenCV~\cite{hirschmuller08} and LIBELAS~\cite{geiger10}. The challenge in this case is due to the fact that the cameras move and the alignment of epipolar lines (image rectification) has to be repeated every time the cameras move. This is achieved by estimating the extrinsic parameters at runtime using SIFT matching and knowledge of the robot kinematics (details are in~\cite{fanello14depth}). For tasks related to object manipulation and visual attention it is enough to compute depth by analyzing pairs of images from the stereo system in isolation. To support motion planning and visual perception for locomotion, however, it is required to merge subsequent maps. Building a consistent map from multiple views requires robust estimation of relative camera motion, fusing together visual matches and robot kinematics (estimated from the motor encoders or the embedded inertial units). To solve this problem we have proposed a framework for robust non linear-least squares which can run in real-time on the robot~\cite{abuhashim16}.

The problem of object perception has been addressed by implementing various techniques based on SIFT features and boosting~\cite{ciliberto11mil}, bag-of-words~\cite{ciliberto13} and, more recently, deep convolutional neural networks~\cite{pasquale2016}. The crucial problem in this context is to study strategies that allow the robot to acquire training examples in real-time either autonomously or with human supervision. In~\cite{fanello2013weakly} we introduced the scenario in which a human teacher shows new objects to the iCub and training data is acquired by segmenting objects using motion~\cite{fanello2013weakly} or disparity~\cite{Pasquale201bfrontiers}. To benchmark the object recognition capabilities of the iCub, we have acquired a dataset which contains images of objects during various training sessions (i.e. the iCub World dataset~\cite{Fanello2013CVPRws, pasquale2016}). We are currently investigating to what extent the performance of the recognition system is affected by the size of the training set (i.e. trade-off in terms of instances of individual objects or samples from different categories), its variability (different illumination conditions, or geometric transformations) and noise (incorrect labels or inaccurate bounding box).

In various work, these capabilities have been exploited to guide the interaction with objects and tools, for example by studying grasping of unknown objects using partial point clouds from stereo-vision~\cite{gori2013,gori2014,vezzani2017} and tool-use~\cite{mar2015,mar2017}.

\begin{figure}[b]
	\centering
	\subfloat[]{\includegraphics[width=0.28\textwidth]{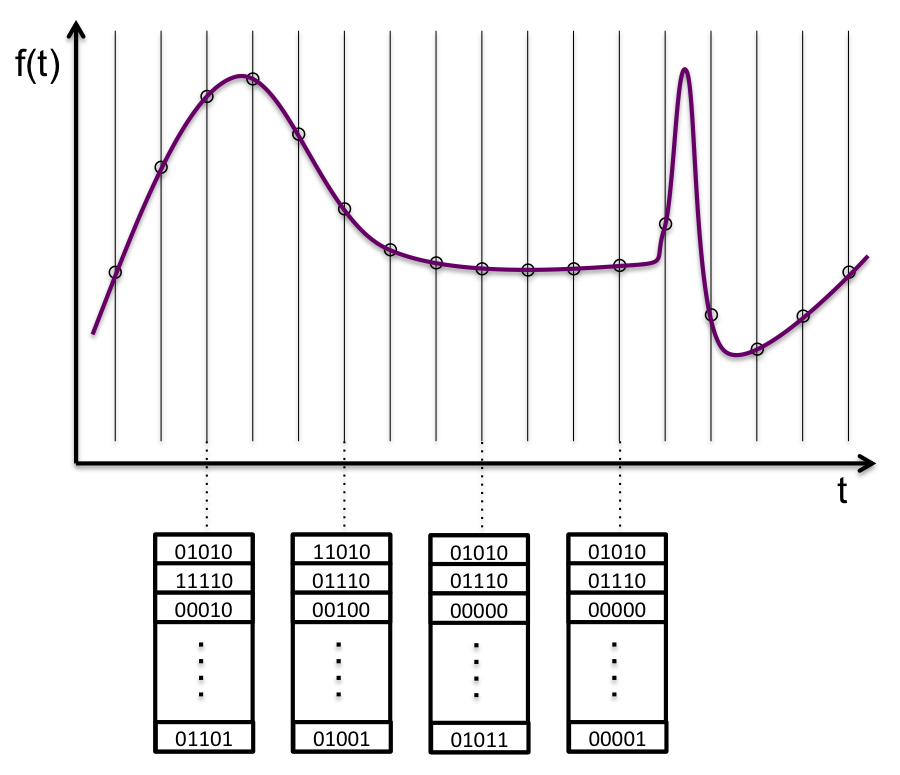}}%
	\subfloat[]{\includegraphics[width=0.28\textwidth]{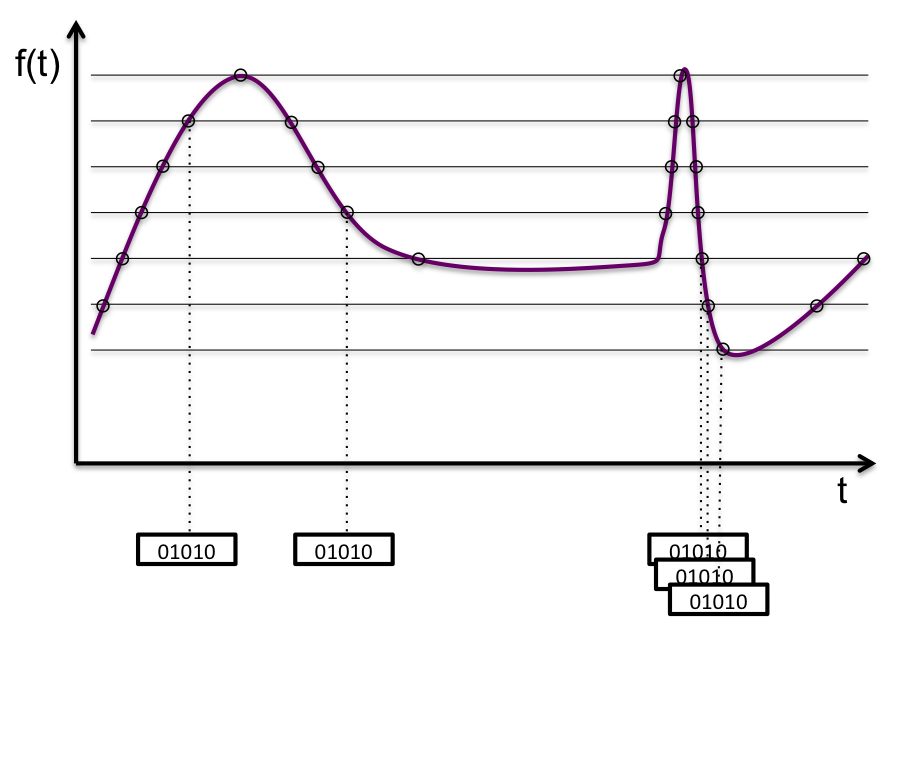}}
	\subfloat[]{\includegraphics[width=0.4\textwidth]{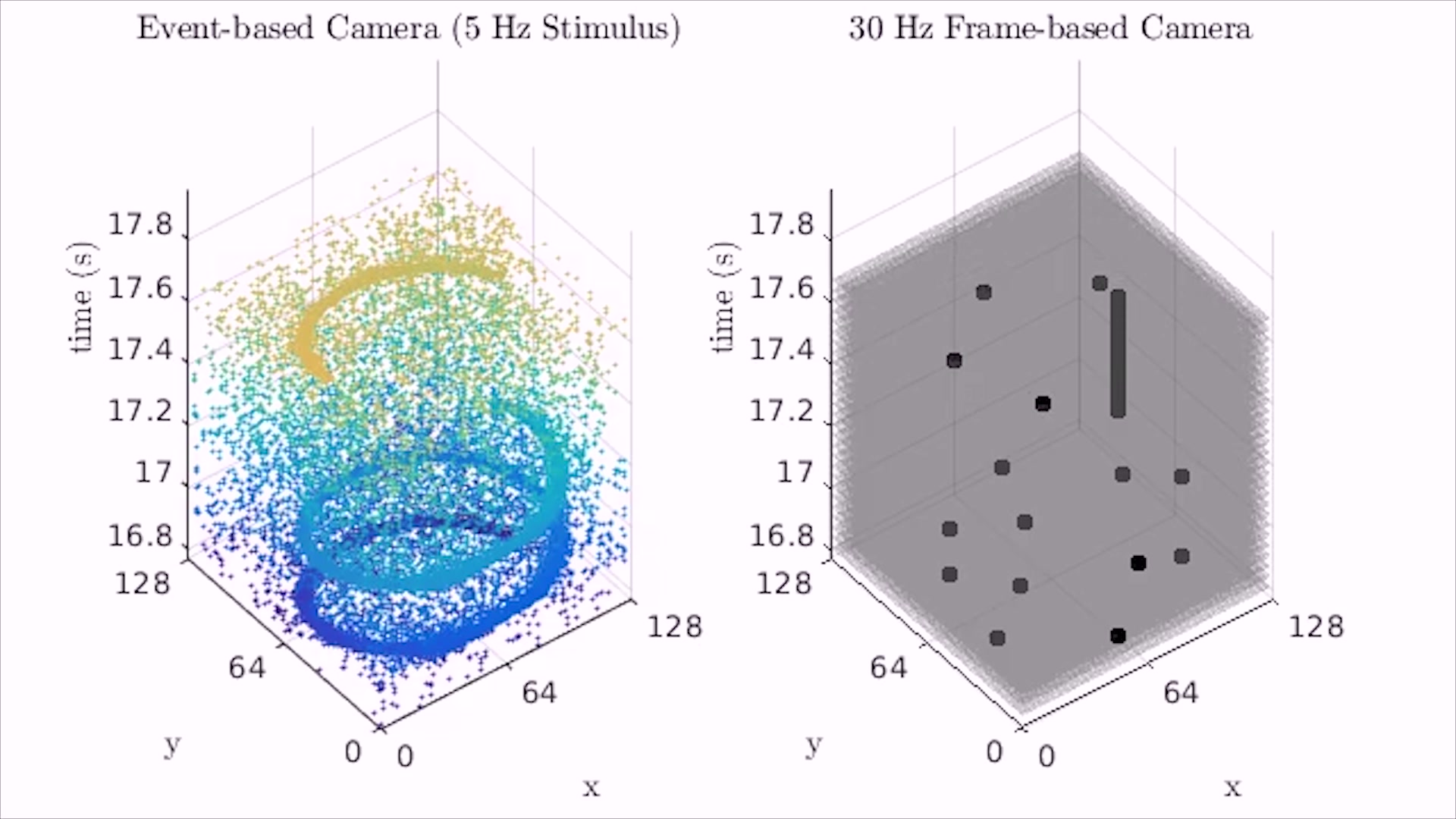}}
	\caption{Event-driven (ED) versus Clock-driven (CD) sensory acquisition: (a) CD sensors sample the sensory signal at fixed times, all of the sensing elements are sampled and the digital value corresponding to the signal amplitude are recorded and sent -- variations above Nyquist rate are lost, while constant signals are oversampled; (b) ED sensors sample the signal over the amplitude dimension, only sensing elements that observe a fixed variation send out a digital pulse, information is in the identity of the sensing element that emitted the pulse and the time at which the pulse has been emitted -- sampling is adaptive, constant singals are not sampled, sampling increases with the derivative of the signal; (c) spatiotemporal activation of ED (left) and CD (right) vision sensors for a black dot rotating in front of the sensor.}
	\label{fig:ed}
\end{figure}
\section{Event-Driven Sensing and Processing}

Efficient encoding of sensory signals allows for an optimal representation of information, reducing cost of acquiring, transmitting and storing unnecessary data, and at the same time, allowing for a better way to extract relevant information. These are important aspects that we are studying towards the development of robots that are energetically and computationally autonomous.

Clock-driven sensing is based on the acquisition of the sensory signals from all of the sensing elements in the sensor (e.g. the pixels in a matrix of an image) at fixed temporal intervals, triggered by a timing signal internally generated that does not depend on the dynamics of the input signal itself (Figure~\ref{fig:ed}). 
Event-driven (ED) sensors, on the other hand, respond to "events" in the sensed signal, usually corresponding to a relative variation of the input~\cite{Liu_Delbruck10}. In this way, the sampling depends on the intensity of the signal itself and intrinsically adapts to the temporal dynamics of the stimulus
, as shown in Figure~\ref{fig:ed}. Redundant information is discarded at the lowest level, increasing the efficiency of the encoding and, hence, of acquisition, communication and processing. At the same time, each sensing element in the sensor array independently sends digital pulses to an external bus as soon as the event is detected. In this configuration, the output of the sensor is an asynchronous stream of digital pulses that encode the information about the sensory signal in the identity (or address) of the sensing element and in the timing between the pulses. The asynchronous communication, not limited by the fixed sampling rate of the whole sensory array, reduces temporal latency of several orders of magnitude, allowing for the implementation of fast sensorimotor loops required for the interaction of robots with highly dynamic environments~\cite{Delbruck_Lang13}.

In the attempt to improve iCub's autonomy, robustness and perceptual skills, one robot has been dedicated to the integration and validation of the event-driven approach. Two Dynamic Vision Sensors~\cite{Lichtsteiner_etal08} replace the Dragonfly2 cameras and are seamlessly integrated in YARP for sensory processing. This unconventional configuration requires the development of novel strategies and algorithms for taking full advantage of the event-driven nature of the sensory system. We started this work by implementing low-level visual modules for the iCub such as event-driven attention, optical flow and learning of spatiotemporal filters, instrumental for object recognition. These applications exploit the low latency, dynamic compression and high temporal resolution of event-driven sensing and processing. Additionally, event-driven sensors convey both spatial and temporal information, e.g. where a stimulus is and its visual appearance, but also when exactly a stimulus moves and how its appearance changes over time. Temporal content of the visual stream is intrinsically neglected in clock-driven systems, its availability in event-driven sensing opens up novel theoretical questions on sensory encoding and computation, that we explored in the context of object recognition with tools of information theory.

\begin{figure}[b]
	\centering
	\subfloat[]{\includegraphics[width=0.3\textwidth]{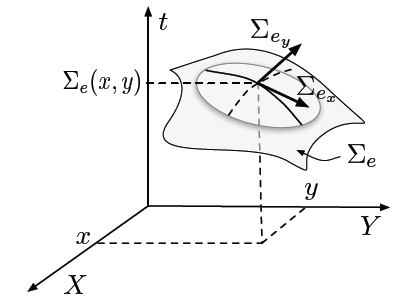}}%
	\subfloat[]{\includegraphics[width=0.3\textwidth]{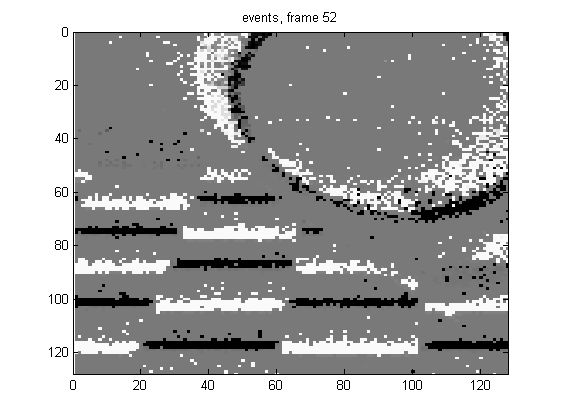}}
	\subfloat[]{\includegraphics[width=0.3\textwidth]{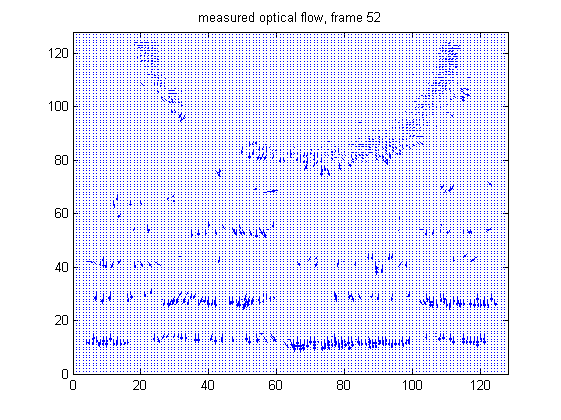}}
	\caption{Optical flow: (a) Surface of events $\Sigma_e$; (b) Events collected over 10ms when the iCub is moving forward (on a mobile platform) and (c) corresponding optical flow. The circular shape is an obstacle, detected as it does not correspond to the expected optical flow. Adapted from~\cite{Benosman_etal14,Clady_etal14}.}
	\label{fig:OpticalFlow}
\end{figure}

\begin{figure}[b]
	\centering
	\subfloat[]{\includegraphics[width=0.33\textwidth]{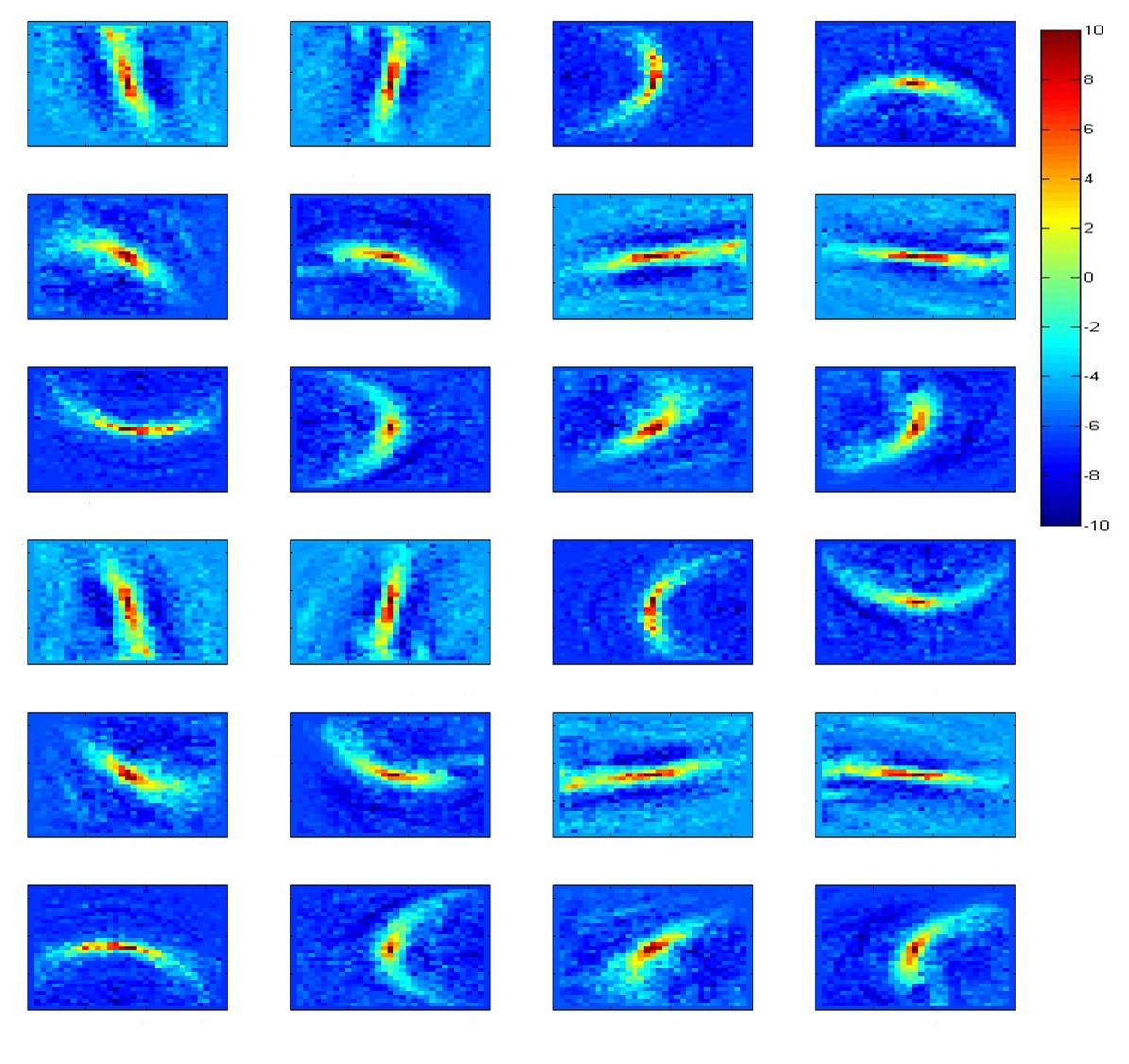}}%
	\subfloat[]{\includegraphics[width=0.5\textwidth]{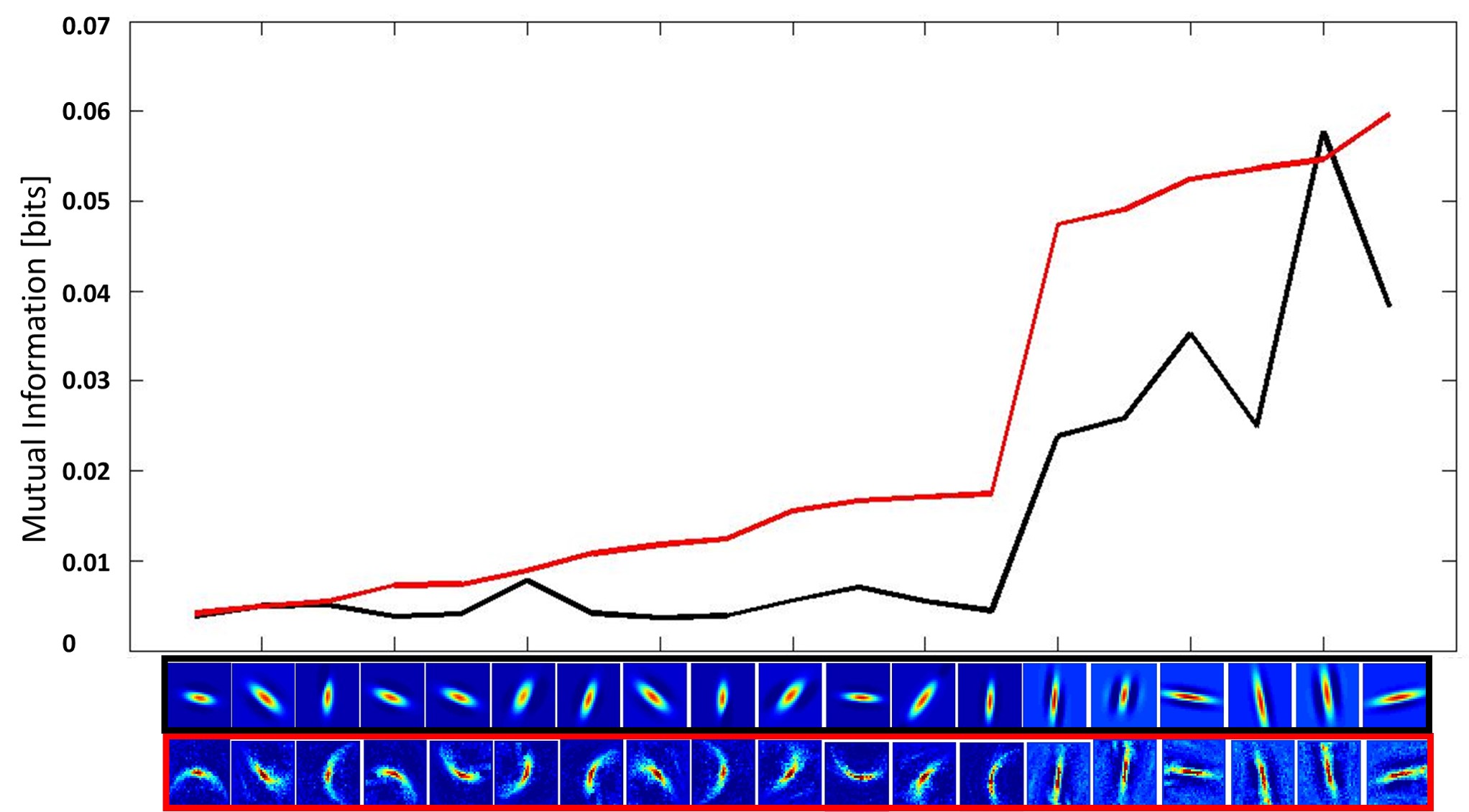}}
	\caption{Spike-timing dependent plasticity, learnt receptive fields: (a) each filter is a $32 \times 32$ matrix in the range $[-10, 10]$. To cover more orientations, the original 6 RF were rotated by $90^o$ four times, generating the total $24$ RFs. (b) Information obtained from each of the learnt receptive fields (red) and their corresponding Gabor filters (black). In abscissa the receptive fields in order of increasing information. Adapted from~\cite{Akolkar_etal15a}.}
	\label{fig:LearntRFs}
\end{figure}

\subsection{Bottom-up attention system and Optical Flow}
Our bottom-up attention system~\cite{Rea_etal13} takes advantage of the intrinsic segmentation and sensitivity to motion that make event-driven sensors useful for fast detection of salient stimuli. The system we have implemented is based on a saliency map model~\cite{Itti_Koch00}: it computes oriented edges and motion from event-driven sensors and it results in a latency of few $ms$ and lower computational load with respect to the correspondent implementation based on clock-driven input (that shows a latency between $50$ and $60~ms$)~\cite{Rea_etal13}.


The temporal information from ED sensors can be used for the real-time computation of the optical flow (Figure~\ref{fig:OpticalFlow}). The almost continuous stream of events gathered from the event-driven cameras creates a monotonically increasing surface $\Sigma_e$ in the $(t,x,y)$ space. The gradient of this surface $\nabla\Sigma_e=(\frac{1}{v_x},\frac{1}{v_y})^T$ measures the rate of change of time with respect to space and its components are the inverse of the velocity at the point where an event has been detected~\cite{Benosman_etal14}. In this implementation, the optical flow is sparse, computed only in the pixels where there is motion, and dense over time, as it is computed for each event, with a temporal resolution of 1$\mu$s, paving the way for fast reactive behavior of robots.

\subsection{Learning Spatiotemporal Filters}

Convolutional Neural Networks (CNN) and Deep Learning architectures are currently the best performing methods for object recognition. Their event-driven implementation has the advantage of performing ``pseudo-simultaneous'' computation, whereby the signals propagate through the network asynchronously and the result of the computation is stable as soon as enough events reach the higher levels of the hierarchy~\cite{Camunas-Mesa_etal14} and almost simultaneous with stimulus presentation.
A way to exploit the temporal information in the stream of events from visual stimuli is to use spike-timing dependent plasticity~\cite{Caporale2008,Meliza2006} to learn spatiotemporal filters for CNN from stimulus statistics~\cite{Akolkar_etal15a}.
The receptive fields learned with this method are similar to traditional Gabor filters, but with a curved shape, as shown in Figure~\ref{fig:LearntRFs}, most importantly, they seem to be able to extract more information from the input stimulus then traditional Gabor filters with similar parameters. Correspondingly, for a very simple feed-forward network, the classification of two stimuli (a ball and a cube) is higher when using the learned filters.


\section{Lessons learned: towards iCub 3.0}
\label{sec:icub30}

The main challenge in making the iCub, a research platform, was to take into consideration requirements deriving from different research lines. We aimed primarily at allowing sophisticated manipulation, using vision and touch, at giving the robot a substantial bi-manual workspace, and eventually a reasonable mobility (although on a tether). We explicitly chose to have only sensors that were approximately ``human'' (e.g. we exlude RGB-D sensors and laser scanners). We wanted to have full programmability at the lowest possible level and, therefore, we designed custom electronics with embedded microcontrollers. In addition, we wanted to make iCub the platform of choice for researchers in artificial cognition (in the broadest sense of the term, i.e. including control, vision, speech and language), which led to the choice of an Open Source approach for both hardware and software.

The mechanical design started from the hands. Since achieving sufficient dexterity with the planned iCub overall size was difficult, we wanted to determine, soon in the design phase, potential difficulties. Many industrial manipulators show a modular design whereby rotary joints of different sizes in series make up the robot. On the iCub, we had to start by designing the hands and allocate enough room in the forearm for the hand’s motors. Fingers move by pulling flexible tendons routed through Bowden’s tubes. There is little space left in the forearm and, consequently, we had to place the elbow motor up in the arm and, eventually, the shoulder motors in the chest. Tendon drives were the only feasible solution to carry movement flexibly from the motors to the joints. We repetitively applied this idea of optimizing the motor position and then routing a set of steel tendons via sets of pulleys. It is fair to say that, although tendon driven joints are a beautiful solution that guarantee a particularly lean design, they are difficult to manage especially for non-experts. This was an important decision point with a complex tradeoff between dependability and the overall shape of the robot.

Once the hand’s design was final, the relative size of all other links and joint ranges was determined from anthropometric tables. To determine joint torques and speeds -- i.e. power and consequently size of the motors -- we ran a number of simulations for several, typical, dynamical tasks. We then crosschecked the simulation results through a simple quasi-static analysis. Finally, we went to an intensive search for motor-gears combinations to achieve the desired torques, size, weight ratios, which gave one single possible candidate for the larger joints: that is, high performance brushless motor coupled with harmonic drive gear. We further had to optimize size by utilizing frameless motors and gears embedding them into the robot’s mechanical structure.

Custom electronics was fundamental to enable tight integration in the available space, to guarantee the flexibility of compensating for joint couplings, manage limits, perform friction compensation, and overall design control laws beyond a simple PID (e.g. force/torque control).

From the hardware point of view, the main challenge in designing (and building) the iCub was on fitting electronics, cables, motors and transmissions in the available space. Design followed an integrated approach with continuous optimization of the each component, but more importantly, taking into account interactions between components. These led to several re-design phases and hundreds of small mechanical debugging actions. Integral to the approach was the availability of an enthusiastic research community, which continuously reported issues in the earlier versions of the robot. The open call of the RobotCub project was instrumental to creating the seed of this community.

\begin{figure}[b]
\centering
\includegraphics[width=10cm]{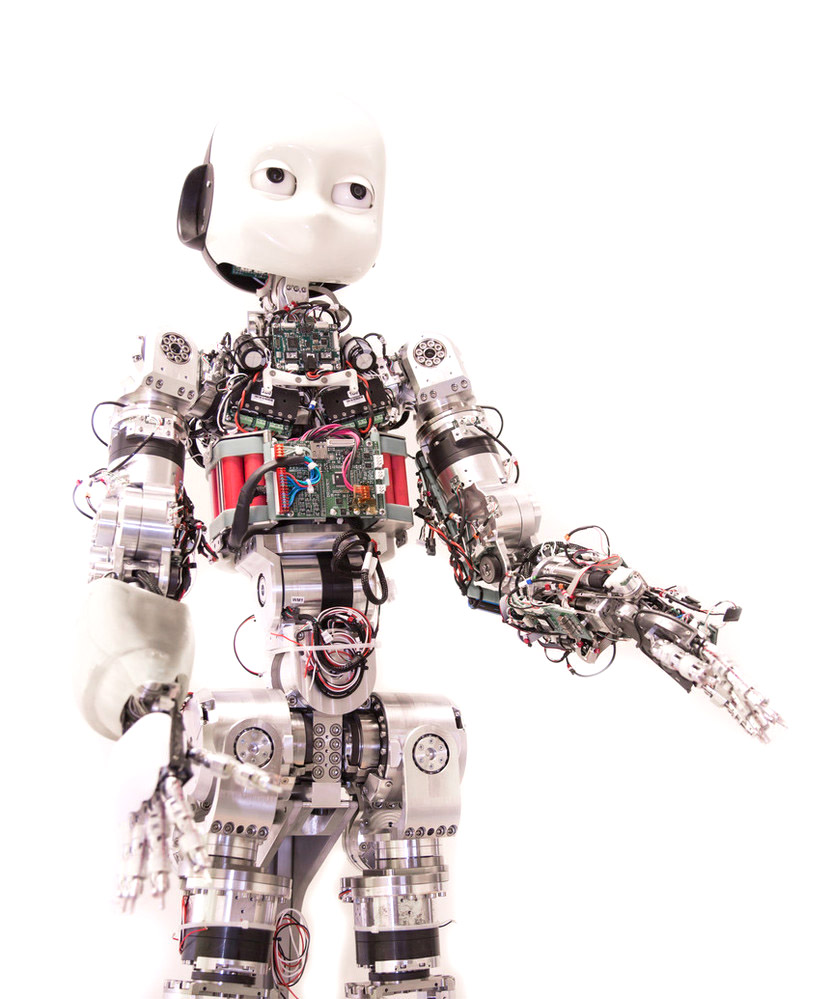}
 \caption{A picture of the iCub 3.0.}
    \label{fig:icub30}
\end{figure}

The initial design of the iCub foresaw tasks related to manipulation and somewhat ``slow'' crawling on all fours. As interests within the research community shifted towards bipedal locomotion, we went for a redesign stage (as described in Section~\ref{sec:icub2}). Simultaneously, we also figured out that we needed more accurate eye/head movements for stereo vision to be useful in guiding the robot’s movements in space. This led to the evolution from iCub version 1.0 to the present mainstream version 2.5, which sports all modifications described above, new Ethernet based electronics as well as the complete skin system (described in Section~\ref{sec:tactile-system}).

In retrospect, we are aware that dealing with the iCub is not always easy. The software shields the user from some of the quirks of the underlying hardware, but problems still creeps in, especially for those who are untrained in robotics. This limits the use of the iCub unless technical support is available. For these reasons, in the new version 3.0, we decided to remove some of the tendon driven solutions. We replaced the differential joints in the shoulder and waist with more traditional serial kinematics that uses a new harmonic drive packaged gear particularly optimized in size. Because of the growing interest in mobile manipulation, the ability to move faster is fundamental. We redesigned the legs to generate twice the amount of power (per joint) of iCub 2.5, enabling the generation of faster walking patterns, longer and more natural steps.

To improve autonomy, iCub 3.0 hosts a 10 Ah battery in the chest and Wi-Fi connectivity. These two features allow it to run without tether. In several joints, we incorporated a safety “clutch” that improves robustness in case of unforeseen impacts as e.g. those due to manipulation tasks~\cite{parmiggiani14}.

To accommodate the above changes, iCub 3.0 resulted 20 cm taller than the iCub 2.5, although the kinematic structure is similar. A first prototype of the iCub 3.0, shown in Figure~\ref{fig:icub30} without the plastic covers and tactile system, is under the final debugging stage.

The iCub development was a worth and exciting endeavor. We believe that the project was largely successful (presently 36 robots worldwide). There can be uncountable factors that helped the iCub project to blossom as e.g. the lack of competitors, the support from the Italian Institute of Technology, the generous funding from the European Commission through its FP6, FP7 and H2020 programs or its physical appearance (small and compact). We can speculate that there may be another reason. iCub is average in many respect, resulting of a wise usage of standard robotic technology. It fits many purposes: it is not the ultimate walking robot but it can walk, it does not possess the most complicated hand but it is dexterous enough. In short, it strikes a balance. The iCub is the compromise of many tradeoffs, but, as a result, it is the only platform where researchers can study vision and touch, whole-body control and manipulation, bimanual manipulation using touch and gesturing, sound localization and speech, language learning as well as human-robot interaction. In other words, there is no other robot like the iCub!

\section{Conclusions}

In the past years the iCub platform has been evolving at a constant pace, both in terms of hardware and software capabilities. In this Chapter we have provided an overview of the historical evolution of the platform. We have started by describing the first version developed in the RobotCub project and the changes introduced in subsequent  revisions, which led to what we called iCub 2.5. We have described the development of the tactile system, which covers the body of the robot and provides contact information for whole-body control and object manipulation. To complement the historical evolution of the robot, in the second part of the Chapter we have made an excursion to illustrate our research in terms of whole-body dynamic control, visual perception and event-driven sensing. These sections should give the reader plenty of references to understand our ongoing effort to progress how iCub perceives and moves in the environment. Finally, in the last section we concluded the paper by discussing some lessons learned and a preview of the changes that will be introduced in the next release, iCub 3.0. So much work has been done, which would be impossible to include here. Nevertheless we hope that this Chapter may serve as a useful reference for researchers that would like to join the iCub community or simply get up to date with the latest developments of the robot and understand how it evolved in the past 12 years.

\begin{acknowledgement}
The authors would like to acknowledge the support of the European Union, which -- since the beginning and through several grants -- has contributed to advancing the iCub platform in all its aspects. In particular we would like to explicitly acknolwedge fundings from the following projects: European FP7 ICT projects No. 611832 (WALK-MAN), No. 600716 (CoDyCo), No 611909 (Koroibot), No 610967 (TACMAN), No 270273 (Xperience) and No 231467 (eMorph).

We would like also to acknowledge the contribution of all researchers and technicians working in the iCub Facility Department: without their skills and dedication not much of what described in the Chapter would have been possible.

\end{acknowledgement}
%

%
%
\bibliographystyle{svmult/styles/spmpsci}
\bibliography{main}

\end{document}